\def\year{2020}\relax
\documentclass[letterpaper]{article}
\usepackage{aaai20}
\usepackage{times}
\usepackage{helvet}
\usepackage{courier}
\usepackage{url}
\usepackage{graphicx}
\usepackage{arydshln}
\usepackage{amssymb}
\usepackage[linesnumbered,ruled]{algorithm2e}
\usepackage{multirow}     
\usepackage{multicol} 
\usepackage{booktabs}
\usepackage{mathtools}

\usepackage{pdfpages}

\newcommand{\citet}[1]
{\citeauthor{#1}~\shortcite{#1}}
\newcommand{\citep}{\cite}

\nocopyright

\DeclareMathOperator*{\EE}{\mathbb{E}}

\title{Causally Denoise Word Embeddings Using Half-Sibling Regression}
\author{
Zekun Yang\thanks{Equal contribution.}\\
Department of Information Systems\\College of Business\\City University of Hong Kong\\Hong Kong SAR, China\\
zekunyang3-c@my.cityu.edu.hk
\\\And
Tianlin Liu\footnotemark[1]\thanks{Corresponding author.}\\
Friedrich Miescher Institute for Biomedical Research\\
Maulbeerstrasse 66\\
4058 Basel, Switzerland\\
tianlin.liu@fmi.ch
}

\begin{document}
\maketitle

\begin{abstract}
Distributional representations of words, also known as word vectors, have become crucial for modern natural language processing tasks due to their wide applications. Recently, a growing body of word vector postprocessing algorithm has emerged, aiming to render off-the-shelf word vectors even stronger. In line with these investigations, we introduce a novel word vector postprocessing scheme under a \emph{causal inference} framework. Concretely, the postprocessing pipeline is realized by Half-Sibling Regression (HSR), which allows us to identify and remove confounding noise contained in word vectors. Compared to previous work, our proposed method has the advantages of interpretability and transparency due to its causal inference grounding. Evaluated on a battery of standard lexical-level evaluation tasks and downstream sentiment analysis tasks, our method reaches state-of-the-art performance.
\end{abstract}

\section{Introduction}

Distributional representations of words have become an indispensable asset in natural language processing (NLP) research due to its wide application in downstream tasks such as parsing \citep{bansal2014tailoring}, named entity recognition \citep{lample2016neural}, and sentiment analysis \citep{tang2014learning}. Of these, ``neural'' word vectors such as Word2Vec \citep{Mikolov2013}, GloVe \citep{Pennington2014}, and Paragram \citep{wieting2015paraphrase} are amongst the most prevalently used and on which we focus in this article. 

There has been a recent thrust in the study of word vector postprocessing methods \citep{Faruqui2015,Fried2015,Mrksic2016,Mrksic2017,Shiue2017,Mu2018,Liu2019,Tang2019}. These methods directly operate on word embeddings and effectively enhance their linguistic regularities in light-weight fashions. Nonetheless, existing postprocessing methods usually come with a few limitations. For example, some rely on external linguistic resources such as English WordNet \citep{Faruqui2015,Fried2015,Mrksic2016,Mrksic2017,Shiue2017}, leaving out-of-database word vectors untouched. Others use heuristic methods to flatten the spectrum of word vector embedding matrices \citep{Mu2018,Liu2019,Wang2019,Tang2019}. Although being effective, these spectral flattening algorithms are primarily motivated by experimental observations but lack of direct interpretability.

In this paper, we propose a novel word vector postprocessing approach that addresses these limitations. Under a \emph{causal inference} framework, the proposed method meets the joint desiderata of (1) \emph{theoretical interpretability}, (2) \emph{empirical effectiveness}, and (3) \emph{computational efficiency}. Concretely, the postprocessing pipeline is realized by Half-Sibling Regression (HSR) \citep{Scholkopf2016}, a method for identifying and removing confounding noise of word vectors. Using a simple linear regression method, we obtain results that are either on-par or outperform state-of-the-art results on a wide battery of lexical-level evaluation tasks and downstream sentiment analysis tasks. More specifically, our contributions are as follows:

 \begin{itemize}
 
 \item We formulate the word vector postprocessing task as a confounding noise identification problem under a putative causal graph. This formulation brings causal interpretability and theoretical support to our postprocessing algorithm.
 
\item The proposed method is data-thrifty and computationally simple. Unlike many existing methods, it does not require external linguistic resources (e.g., synonym relationships); besides, the method can be implemented easily via simple linear regressions.

\item The proposed postprocessing method yields highly competitive empirical results. For example, while achieving the best performance on 20 semantic textual similarity tasks, on average, our proposed method brings 4.71\%, 7.54\%, and 6.54\% improvement respectively compared to the previously best results, and it achieves 7.13\%, 22.06\%, and 9.83\% improvement compared to the original word embedding when testing on Word2Vec, GloVe, and Paragram.
\end{itemize} 
 
The rest of the paper is organized as follows. We first briefly review prior work on word vector postprocessing. Next, we introduce Half-Sibling Regression as a causal inference framework to remove confounding noise; we then proceed to explain how to apply Half-Sibling Regression to remove noise from word embeddings. Then, we showcase the effectiveness of the Half-Sibling Ridge Regression model on word similarity tasks, semantic textual similarity tasks, and downstream sentiment analysis tasks using three different pre-trained English word embeddings. Finally, we conduct statistical significance tests on all experimental results\footnote{Our codes are available at \url{https://github.com/KunkunYang/denoiseHSR-AAAI}}.
 
\section{Prior Work} \label{sec:prior}
 In this section, we review prior art for word vector postprocessing. Modern word vector postprocessing methods can be broadly divided into two streams: (1) lexical and (2) spatial approaches.

\paragraph{The Lexical Approach} The lexical approach uses lexical relational resources to enhance the quality of word vectors. These lexical relational resources specify semantic relationships of words such as synonym and antonym relationships. For example, \citet{Faruqui2015} inject synonym lexical information into pre-trained collections of word vectors. \citet{Mrksic2016} generalize this approach and insert both antonym and synonymy constraints into word vectors. \citet{Mrksic2017} use constraints from mono- and cross-lingual lexical resources to fine-tune word vectors. \citet{Fried2015} and \citet{Shiue2017} propose to use hierarchical semantic relations such as hypernym semantics to enrich word vectors. To make sure that word vectors satisfy the lexical relational constraints, supervised machine learning algorithms are used.

\paragraph{The Spatial Approach} The spatial approach differs from the lexical approach in that it does not require external knowledge bases. The general principle of this approach is to enforce word vectors to be more ``isotropic'', i.e., more spread out in space. This goal is usually achieved by flattening the spectrum of word vectors. For example, \citet{Mu2018} propose All-But-The-Top (ABTT) method which removes leading principal components of word vectors; \citet{Wang2019} extend this idea by softly shrinking principal components of word embedding matrix using a variance normalization method; \citet{Liu2019} propose the Conceptor Negation (CN) method, which employs regularized identity maps to filter away high-variance latent features of word vectors;  more recently, \citet{Tang2019} develop SearchBeta (SB) that uses a centralized kernel alignment method to smooth the spectrum of word vectors.

\section{The Causal Inference Approach for Word Vector Postprocessing}

The lexical and spatial approaches introduced in the previous section have empirically proven to be effective. Nonetheless, they also suffer from a few limitations. A shortcoming of the lexical approach is that it is unable to postprocess out-of-database word vectors. Indeed, lexical relational resources like synonym-antonym relationships are informative for word meaning, in particular word meaning of \emph{adjectives}. However, many non-adjective words do not have abundant lexical connections with other words, and for this reason, they are not well-represented in lexical-relationship databases. For instance, most nouns (e.g., \texttt{car}) and verbs (e.g., \texttt{write}) have few synonyms and even fewer antonyms, making the lexical postprocessing methods inapplicable to these words. The spatial approach favorably avoids this problem by lifting the requirement of lexical relational resources. Yet, one major downside of the spatial approach is its lack of direct interpretability. For example, many spatial approaches propose to completely or softly remove a few leading principal components (PCs) of word vectors. However, it is rather unclear what exactly has been encoded by these leading PCs other than the empirical finding that these leading PCs are somehow correlated with word frequencies \citep{Mu2018}.

In this paper, we go beyond the lexical and spatial schemes and introduce a novel \emph{causal inference approach} for postprocessing word vectors. The method does not seek to infer the causal structure of words or word vectors; instead, in line with \citet{Scholkopf2012On} and \citet{Scholkopf2016}, it incorporates causal beliefs and assumptions for empirical objectives -- postprocessing off-the-shelf word vectors in our case. Concretely, this is achieved by identifying and removing confounding noise of word vectors using Half-Sibling Regression (HSR) method \citep{Scholkopf2016}. Here we first briefly introduce HSR and then explain how to apply HSR to word vectors.

\subsection{Half-Sibling Regression}

\begin{figure}[ht]
\centering
\includegraphics[width = 0.4\textwidth]{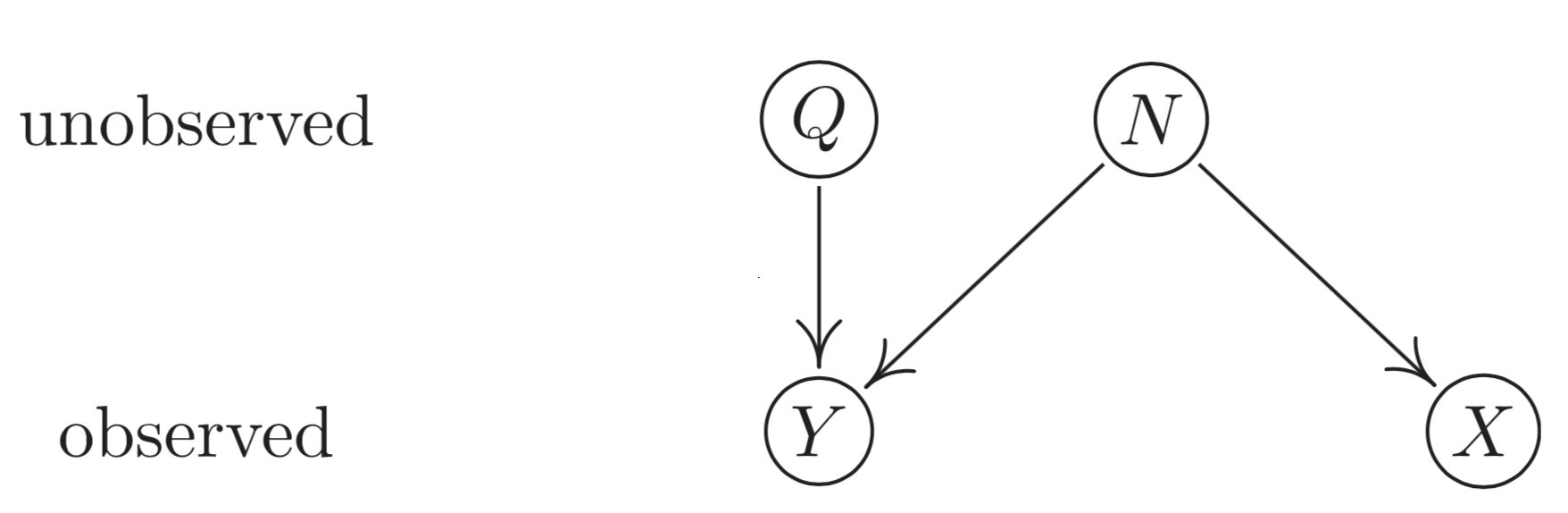}
\caption{The causal graph for HSR (adapted from \citet{Scholkopf2016}). Each vertex labeled by $Q$, $N$, $Y$, and $X$ is a random variable defined on an underlying probability space. Directed edges connecting random variables describe probabilistic dependency between random variables.}
\label{fig:causalGraph}
\end{figure}

In the passing, we introduce HSR mainly based on the presentation of \citet{Scholkopf2016}. Consider a hypothetical causal graph, shown in Figure \ref{fig:causalGraph}, where each vertex labeled by $Q$, $N$, $Y$, and $X$ are random variables defined on an underlying probability space and each directed edge indicates the probabilistic dependency between two random variables. We are mostly interested in quantities taken by the random variable $Q$. Unfortunately, it is not possible to directly observe these quantities. Instead, we are given only the \emph{corrupted} observations of $Q$, taken value by the random variable $Y$. That is, intuitively $Y$ can be seen as a noisy, lossy version of $Q$. A natural assumption of $Y$ is that it statistically depends on its ``clean'' version $Q$ as well as some noise, whose values are taken by some unobservable random variable $N$ that encodes the noise source. We further assume that the noise source $N$ affects another random variable, $X$, whose quantities are directly observable. Importantly, we require $X$ to be independent of $Q$.

Recall that we are mostly interested in the unobservable random variable $Q$. Hence the question we aim to answer is: How to reconstruct the quantities taken by $Q$ by leveraging the underlying statistical dependency structure in Figure \ref{fig:causalGraph}? HSR provides a simple yet effective solution to this question -- It estimates $Q$ via its approximation $\hat{Q}$, which is defined as

\begin{equation} \label{eq:HSR}
\hat{Q} \coloneqq Y - \EE [Y \mid X].
\end{equation}

The HSR Equation \ref{eq:HSR} can be straightforwardly interpreted as follows. Recall that $X$ is independent of $Q$, and therefore $X$ is \emph{not} predictive to $Q$ or $Q$'s influence on $Y$. However, $X$ is predictive to $Y$, because $X$ and $Y$ are both influenced by the \emph{same} noise source $N$. When predicting $Y$ based on $X$ realized by the term $\EE [Y \mid X]$, since those signals of $Y$ coming from $Q$ cannot be predicted by $X$, only those noise contained in $Y$ coming from $N$ could be captured. To reconstruct $Q$ from $Y$, we can therefore remove the captured noise $\EE [Y \mid X]$ from $Y$, resulting in the reconstruction $\hat{Q} \coloneqq Y - \EE [Y \mid X]$, which is Equation \ref{eq:HSR}. This procedure is referred to as Half-Sibling Regression because $X$ and $Y$ share one parent vertex $N$. $Y$ is regressed upon its half-sibling $X$ to capture the components of $Y$ inherited from their shared parent vertex $N$.

HSR enjoys a few appealing theoretical properties. In particular, it is possible to show that $\hat{Q}$ reconstructs $Q$ (up to its expectation $\EE[Q]$) at least as good as the mean-subtraction $Y - \EE [Y]$ does. We refer the readers to \citet{Scholkopf2016} for detailed theoretical discussions.

\subsection{Applying HSR to De-Noise Word Vectors}

We now explain how we apply HSR to remove noise from word vectors. Before getting into the details, we first recall some linguistic basics of words, which are the key enablers of our approach. Semantically, words can be divided into two basic classes: (1) content or open-class words and (2) function or closed-class words (also known as stop words). Content words are those that have meaning or semantic value, such as nouns, verbs, adjectives, and adverbs. Function words have little lexical meaning; rather, they mainly exist to explain grammatical or structural relationships with other words. In English, examples of function words include \texttt{a}, \texttt{to}, \texttt{for}, \texttt{of}, \texttt{the}, and more.

\begin{algorithm}[ht]
\SetKwInOut{Input}{Input}
\SetKwInOut{Output}{Output}
\Input{(i) $\{v_i^Y\}_{i = 1}^K$: a collection of $K$ content-word vectors, each of dimension $n$; $\mathbf{V}^Y$ is a $n \times K$ matrix whose columns are from $\{v_i^Y\}_{i = 1}^K$. (ii) $\{v_i^X\}_{i = 1}^P$: a collection of $P$ function-word vectors, each of dimension $n$; $\mathbf{V}^X$ is a $n \times P$ matrix whose columns are from $\{v_i^X\}_{i = 1}^P$. (iii) Regression constants $\alpha_1, \alpha_2 > 0$.} 
\textbf{Postprocess content-word vectors}: \newline
 Step 1.1: \emph{Identify noise contained in content-word vectors}: Estimate a weight matrix $\textbf{W}_1$ such that
\[ \mathbf{V}^Y \approx \mathbf{V}^X \mathbf{W}_1, \]
 with ridge regression
\[ \mathbf{W}_1 = \left  ( (\mathbf{V}^X)^\top \mathbf{V}^X  + \alpha_1  \mathbf{I} \right )^{-1} (\mathbf{V}^X)^\top \mathbf{V}^Y.
\]
 Step 1.2: \emph{Remove noise contained in content-word vectors}: 
\[ \hat{\mathbf{V}}^Y \coloneqq \mathbf{V}^Y  - \mathbf{V}^X \mathbf{W}_1. \]  \\

\textbf{Postprocess stop-word vectors}: \newline
 Step 2.1: \emph{Identify noise contained in stop-word vectors}: Estimate a weight matrix $\textbf{W}_2$ such that
\[ \mathbf{V}^X \approx \mathbf{V}^Y \mathbf{W}_2, \]
 with ridge regression
\[ \mathbf{W}_2 = \left  ((\mathbf{V}^Y)^\top \mathbf{V}^Y  + \alpha_2  \mathbf{I} \right )^{-1} (\mathbf{V}^Y)^\top \mathbf{V}^X.
\]
 Step 2.2: \emph{Remove noise contained in stop-word vectors}: 
\[ \hat{\mathbf{V}}^X \coloneqq \mathbf{V}^X  - \mathbf{V}^Y \mathbf{W}_2. \]  \\
\Output{(i) HSR postprocessed content-word vectors $\{\hat{v}_i^Y\}$, which are columns of $\hat{\mathbf{V}}^Y$; (ii) HSR postprocessed stop-word vectors $\{\hat{v}_i^X\}$, which are columns of  $\hat{\mathbf{V}}^X$.}  
\caption{HSR algorithm for word vector postprocessing}
\label{alg:hsr}
\end{algorithm}

Based on these basic linguistic facts, we posit that content-word vectors and function-word vectors can be seen as half-siblings as their linguistic properties align well with the HSR foundations. Indeed, since function-word vectors carry little semantic content, they could not be predictive to clean content-word vectors. Additionally, since content-word vectors and function-word vectors are induced from some shared training corpora, we hypothesize that they are subjected to the same noise profile. Using HSR language of Figure \ref{fig:causalGraph}, this means we can model the off-the-shelf stop-word vectors with $X$, off-the-shelf content-word vectors with $Y$, and ``clean'' yet unseen content-word vectors with $Q$. Under the HSR framework, when we regress content-word vectors upon function-word vectors, only the noise of the former is captured. Once such noises are identified, they can be directly subtracted, so that the clean content-word vectors will be reconstructed.

The above described procedure can be mathematically realized as follows. Let $\{v^X_i\}_{i=1}^P$ be a collection of function-word vectors and let $\{v^Y_i\}_{i=1}^K$ be a collection of content-word vectors. To postprocess content-word vectors $\{v^Y_i\}_{i=1}^K$, we run a simple two-step algorithm. In the first step, we estimate parameters of a linear multiple-output model \citep[Section 3.2.4]{Hastie2001}, in which we use model inputs $v^X_1, \cdots, v^X_P$ to predict model outputs $v^Y_1, \cdots, v^Y_K$. This amounts to estimate each $w_{ij}$ such that $v^Y_j \approx \sum_{i = 1}^P w_{ij} v^X_i$ for each $j \in \{1, \cdots, K\}$. In the second step, we remove the regression result from the target of the regression. That is, we let $\hat{v}^Y_j \coloneqq v_j^Y - \sum_{i = 1}^P w_{ij} v^X_i$ be the postprocessed content-word vector.

So far, we have described how to postprocess content-word vectors. To postprocess function-word vectors, we can employ a similar pipeline with the predictor and target flipped. That is, to identify confounding noise contained in stop-word vectors, we use off-the-shelf content-word vectors as features to predict off-the-shelf stop-word vectors. The full algorithm is summarized in Algorithm \ref{alg:hsr}.

We provide a few remarks on the practical implementations and further generalizations of Algorithm \ref{alg:hsr}. Our first remark goes to how to identify the function and content words in practice. Throughout our experiments, to identify function words, we use the stop word list provided by Natural Language Toolkit (NLTK) package\footnote{\url{https://www.nltk.org/}}, which is a list of 179 words. We regard words outside of this list to be content words. A small amount of stop words works efficiently when postprocessing tens of thousands of content-word vectors because in this case, we only have a handful of features. However, when postprocessing stop-word vectors, it is cumbersome because the number of content words as features are too large to be efficiently implemented. For this reason, in practice, we only use a small sample of commonly used content-word vectors as features for postprocessing stop-word vectors. Specifically, borrowing the word list provided by \citet{Arora2017}, we use the most frequent 1000 content words as features in Step 2.1 and Step 2.2 of Algorithm \ref{alg:hsr}.

Moreover, while our framework postprocesses both content and function words, we have tried only postprocessing content words and leaving function words unchanged. We discover that the experimental results are still better than the baseline spatial approaches but worse than postprocessing both content and function words. The reason might be that stop words play non-trivial roles in various NLP tasks. As all baseline spatial approaches postprocess both content and function words, we follow this setting.

Finally, we remark that the linear model used in Algorithm \ref{alg:hsr} can be straightforwardly generalized to non-linear models. For this, we have formulated and tested Multilayer Perceptrons (MLPs) as extensions to the linear model used in Algorithm \ref{alg:hsr}. The detailed MLP version of Algorithm \ref{alg:hsr} is postponed to the appendix.

\begin{table*}[t]
  \centering
  \caption{Spearman's rank correlation coefficient of seven word similarity tasks}
\scalebox{0.72}{

             \begin{tabular}{lrrrrrrrrrrrrrrr}
             \toprule
    \multirow{2}[0]{*}{} & \multicolumn{5}{c}{\textbf{WORD2VEC}} & \multicolumn{5}{c}{\textbf{GLOVE}}    & \multicolumn{5}{c}{\textbf{PARAGRAM}} \\
    \cmidrule(r){2-6} \cmidrule(r){7-11} \cmidrule(r){12-16}
          & \multicolumn{1}{c}{\textbf{Orig.}} & \multicolumn{1}{c}{\textbf{ABTT}} & \multicolumn{1}{c}{\textbf{CN}} & \multicolumn{1}{c}{\textbf{SB}} & \multicolumn{1}{c}{\textbf{HSR-RR}} & \multicolumn{1}{c}{\textbf{Orig.}} & \multicolumn{1}{c}{\textbf{ABTT}} & \multicolumn{1}{c}{\textbf{CN}} & \multicolumn{1}{c}{\textbf{SB}} & \multicolumn{1}{c}{\textbf{HSR-RR}} & \multicolumn{1}{c}{\textbf{Orig.}} & \multicolumn{1}{c}{\textbf{ABTT}} & \multicolumn{1}{c}{\textbf{CN}} & \multicolumn{1}{c}{\textbf{SB}} & \multicolumn{1}{c}{\textbf{HSR-RR}} \\
            \toprule
    \textbf{RG65} & 0.7494 & \underline{0.7869} & \underline{\textbf{0.8041}} & \underline{0.7964} & 0.7569 & 0.7603 & 0.7648 & \underline{\textbf{0.7913}} & \underline{0.7850} & \underline{0.7694} & 0.7630 & \underline{0.7683} & 0.7594 & \underline{\textbf{0.7898}} & \underline{0.7760} \\
    \textbf{WordSim-353} & \underline{0.6999} & 0.6929 & \underline{0.6992} & 0.6856 & \underline{\textbf{0.7059}} & 0.7379 & \underline{0.7668} & \underline{0.7886} & 0.7115 & \underline{\textbf{0.7887}} & 0.7302 & \underline{\textbf{0.7386}} & \underline{0.7321} & 0.7196 & \underline{0.7338} \\
    \textbf{RW} & 0.5997 & 0.5984 & \underline{\textbf{0.6036}} & \underline{0.5998} & \underline{0.6033} & 0.5101 & \underline{0.5716} & \underline{\textbf{0.5898}} & 0.4879 & \underline{0.5580} & 0.5972 & \underline{\textbf{0.6038}} & \underline{0.6006} & 0.5769 & \underline{0.6023} \\
    \textbf{MEN} & 0.7706 & \underline{\textbf{0.7929}} & \underline{0.7901} & \underline{0.7888} & 0.7726 & 0.8013 & \underline{0.8234} & \underline{\textbf{0.8339}} & 0.7853 & \underline{0.8258} & \underline{0.7728} & 0.7705 & \underline{0.7746} & 0.7693 & \underline{\textbf{0.7750}} \\
    \textbf{MTurk} & \underline{0.6831} & 0.6538 & 0.6610 & \underline{0.6846} & \underline{\textbf{0.6854}} & 0.6916 & \underline{\textbf{0.7233}} & \underline{0.7116} & 0.6731 & \underline{0.7074} & \underline{0.6300} & 0.6106 & \underline{0.6251} & 0.6147 & \underline{\textbf{0.6319}} \\
    \textbf{SimLex-999} & 0.4427 & 0.4629 & \underline{\textbf{0.4728}} & \underline{0.4702} & \underline{0.4672} & 0.4076 & \underline{0.4650} & \underline{\textbf{0.4858}} & 0.3985 & \underline{0.4728} & 0.6847 & \underline{0.6862} & 0.6854 & \underline{0.6878} & \underline{\textbf{0.6903}} \\
    \textbf{SimVerb-3500} & 0.3659 & 0.3792 & \underline{0.3868} & \underline{0.3865} & \underline{\textbf{0.3978}} & 0.2842 & \underline{0.3433} & \underline{0.3632} & 0.2671 & \underline{\textbf{0.3980}} & 0.5411 & \underline{0.5461} & \underline{0.5413} & 0.5389 & \underline{\textbf{0.5518}} \\
      \bottomrule
    \end{tabular}
}%
  \label{tab:word_sim}%
\end{table*}%

\begin{table*}[htbp]
  \centering
  \caption{Pearson correlation coefficient of 20 semantic textual similarity tasks}
\scalebox{0.70}{%

   \begin{tabular}{lrrrrrrrrrrrrrrr}
   \toprule
    \multirow{2}[0]{*}{} & \multicolumn{5}{c}{\textbf{WORD2VEC}} & \multicolumn{5}{c}{\textbf{GLOVE}}    & \multicolumn{5}{c}{\textbf{PARAGRAM}} \\
    \cmidrule(r){2-6} \cmidrule(r){7-11} \cmidrule(r){12-16}
          & \multicolumn{1}{c}{\textbf{Orig.}} & \multicolumn{1}{c}{\textbf{ABTT}} & \multicolumn{1}{c}{\textbf{CN}} & \multicolumn{1}{c}{\textbf{SB}} & \multicolumn{1}{c}{\textbf{HSR-RR}} & \multicolumn{1}{c}{\textbf{Orig.}} & \multicolumn{1}{c}{\textbf{ABTT}} & \multicolumn{1}{c}{\textbf{CN}} & \multicolumn{1}{c}{\textbf{SB}} & \multicolumn{1}{c}{\textbf{HSR-RR}} & \multicolumn{1}{c}{\textbf{Orig.}} & \multicolumn{1}{c}{\textbf{ABTT}} & \multicolumn{1}{c}{\textbf{CN}} & \multicolumn{1}{c}{\textbf{SB}} & \multicolumn{1}{c}{\textbf{HSR-RR}} \\
     \toprule
    \textbf{STS-2012-MSRpar} & \textbf{41.78} & 38.70 & 39.42 & 40.77 & 34.42 & \textbf{42.06} & 41.41 & 41.27 & 41.15 & 32.49 & 39.32 & 38.84 & 39.84 & 37.72 & \textbf{41.44} \\
    \textbf{STS-2012-MSRvid} & 76.27 & 75.60 & 75.32 & 74.98 & \textbf{79.63} & 65.85 & 67.84 & 62.50 & 64.71 & \textbf{80.03} & 56.34 & 57.65 & 56.78 & 55.55 & \textbf{62.31} \\
    \textbf{STS-2012-surprise.OnWN} & 70.62 & 70.89 & 70.73 & 69.99 & \textbf{71.27} & 60.74 & 69.48 & 67.87 & 57.02 & \textbf{72.24} & 62.60 & 64.61 & 63.21 & 60.68 & \textbf{67.91} \\
    \textbf{STS-2012-SMTeuroparl} & 31.20 & 35.71 & 35.29 & 33.88 & \textbf{40.32} & 51.97 & \textbf{54.36} & 52.58 & 50.06 & 51.60 & 50.64 & 51.64 & 50.63 & 51.34 & \textbf{51.92} \\
    \textbf{STS-2012-surprise.SMTnews} & \textbf{51.07} & 46.24 & 47.34 & 47.10 & 50.09 & 46.35 & 48.19 & 47.69 & 45.18 & \textbf{54.41} & 52.94 & 50.18 & 52.66 & \textbf{54.16} & 53.87 \\
    \hdashline
    \textbf{STS-2012} & 54.19 & 53.43 & 53.62 & 53.34 & \textbf{55.15} & 53.39 & 56.26 & 54.38 & 51.62 & \textbf{58.15} & 52.37 & 52.58 & 52.62 & 51.89 & \textbf{55.49} \\
    \toprule
    \textbf{STS-2013-FNWN} & 39.68 & 43.51 & 43.40 & 42.95 & \textbf{49.09} & 39.48 & 45.81 & 42.03 & 39.15 & \textbf{46.47} & 35.79 & 36.05 & 35.93 & 34.35 & \textbf{38.00} \\
    \textbf{STS-2013-OnWN} & 67.98 & 70.56 & 69.29 & 69.12 & \textbf{75.57} & 53.75 & 63.86 & 57.45 & 52.36 & \textbf{74.91} & 48.07 & 48.18 & 48.23 & 48.28 & \textbf{56.57} \\
    \textbf{STS-2013-headlines} & 63.29 & 63.24 & 63.62 & 63.22 & \textbf{63.65} & 63.54 & 66.70 & 67.00 & 60.65 & \textbf{68.56} & 64.43 & 65.13 & 64.69 & 62.99 & \textbf{66.90} \\
    \hdashline
    \textbf{STS-2013} & 56.98 & 59.10 & 58.77 & 58.43 & \textbf{62.77} & 52.26 & 58.79 & 55.49 & 50.72 & \textbf{63.31} & 49.43 & 49.79 & 49.62 & 48.54 & \textbf{53.82} \\
    \toprule
    \textbf{STS-2014-OnWN} & 74.85 & 75.92 & 75.27 & 74.43 & \textbf{81.40} & 61.91 & 70.93 & 66.43 & 60.36 & \textbf{81.39} & 60.29 & 61.95 & 60.75 & 59.45 & \textbf{68.30} \\
    \textbf{STS-2014-deft-forum} & 41.30 & 42.25 & 42.74 & 42.03 & \textbf{46.73} & 28.82 & 38.90 & 37.57 & 25.91 & \textbf{45.85} & 35.17 & 37.60 & 35.75 & 33.59 & \textbf{40.84} \\
    \textbf{STS-2014-deft-news} & 66.76 & 64.87 & 65.45 & 64.97 & \textbf{67.88} & 63.41 & 68.72 & 69.08 & 61.27 & \textbf{70.60} & 62.19 & 63.73 & 62.75 & 61.09 & \textbf{66.66} \\
    \textbf{STS-2014-headlines} & 60.87 & 60.61 & \textbf{61.09} & 60.66 & 60.93 & 59.28 & 61.34 & 61.71 & 56.25 & \textbf{64.01} & 60.84 & 60.72 & 60.97 & 60.21 & \textbf{62.83} \\
    \textbf{STS-2014-tweet-news} & 73.33 & 75.13 & 74.87 & 73.66 & \textbf{76.00} & 62.43 & 74.62 & \textbf{75.38} & 58.70 & 75.09 & 69.29 & 72.43 & 70.14 & 66.75 & \textbf{75.16} \\
    \textbf{STS-2014-images} & 77.44 & 77.81 & 78.42 & 77.11 & \textbf{80.55} & 61.89 & 69.40 & 65.81 & 59.03 & \textbf{78.45} & 53.67 & 58.29 & 54.86 & 51.58 & \textbf{65.10} \\
    \hdashline
    \textbf{STS-2014} & 65.76 & 66.10 & 66.31 & 65.48 & \textbf{68.92} & 56.29 & 63.99 & 62.66 & 53.59 & \textbf{69.23} & 56.91 & 59.12 & 57.54 & 55.45 & \textbf{63.15} \\
    \toprule
    \textbf{STS-2015-answers-forums} & 52.65 & 54.01 & 53.99 & 50.51 & \textbf{66.77} & 36.86 & 49.58 & 48.62 & 36.76 & \textbf{65.46} & 38.79 & 41.19 & 39.25 & 38.35 & \textbf{48.37} \\
    \textbf{STS-2015-answers-students} & 70.82 & 70.92 & 71.65 & 69.74 & \textbf{72.16} & 62.77 & 69.46 & \textbf{69.68} & 61.84 & 67.38 & 67.52 & 69.46 & 67.96 & 66.80 & \textbf{71.98} \\
    \textbf{STS-2015-belief} & 60.11 & 61.91 & 61.62 & 58.10 & \textbf{77.08} & 44.20 & 61.43 & 59.77 & 41.19 & \textbf{76.12} & 49.77 & 55.57 & 50.79 & 46.98 & \textbf{61.32} \\
    \textbf{STS-2015-headlines} & 68.11 & 68.28 & 68.65 & 68.19 & \textbf{69.02} & 65.42 & 68.90 & 69.20 & 63.25 & \textbf{71.41} & 67.85 & 68.40 & 68.09 & 66.92 & \textbf{70.38} \\
    \textbf{STS-2015-images} & 80.07 & 80.18 & 80.74 & 79.48 & \textbf{83.08} & 69.14 & 73.53 & 71.43 & 67.81 & \textbf{80.58} & 66.55 & 68.29 & 67.08 & 65.55 & \textbf{73.17} \\
    \hdashline
    \textbf{STS-2015} & 66.35 & 67.06 & 67.33 & 65.20 & \textbf{73.62} & 55.68 & 64.58 & 63.74 & 54.17 & \textbf{72.19} & 58.10 & 60.58 & 58.63 & 56.92 & \textbf{65.04} \\
    \toprule
    \textbf{SICK} & 72.25 & \textbf{72.49} & 72.40 & 72.32 & 72.02 & 66.64 & 68.12 & 66.42 & 66.03 & \textbf{71.62} & 64.55 & 64.89 & 64.78 & 64.05 & \textbf{67.07} \\
    \bottomrule
    \end{tabular}%

    }
  \label{tab:STS}%
\end{table*}%

\begin{table*}[htbp]
  \centering
  \caption{Five-fold cross-validation accuracy of four sentiment analysis tasks}
\scalebox{0.75}{%

    \begin{tabular}{lrrrrrrrrrrrrrrr}
    \toprule
          & \multicolumn{5}{c}{\textbf{WORD2VEC}} & \multicolumn{5}{c}{\textbf{GLOVE}}    & \multicolumn{5}{c}{\textbf{PARAGRAM}} \\
          \cmidrule(r){2-6} \cmidrule(r){7-11} \cmidrule(r){12-16}
          & \multicolumn{1}{l}{\textbf{Orig.}} & \multicolumn{1}{l}{\textbf{CN}} & \multicolumn{1}{l}{\textbf{ABTT}} & \multicolumn{1}{l}{\textbf{SB}} & \multicolumn{1}{l}{\textbf{HSR-RR}} & \multicolumn{1}{l}{\textbf{Orig.}} & \multicolumn{1}{l}{\textbf{CN}} & \multicolumn{1}{l}{\textbf{ABTT}} & \multicolumn{1}{l}{\textbf{SB}} & \multicolumn{1}{l}{\textbf{HSR-RR}} & \multicolumn{1}{l}{\textbf{Orig.}} & \multicolumn{1}{l}{\textbf{CN}} & \multicolumn{1}{l}{\textbf{ABTT}} & \multicolumn{1}{l}{\textbf{SB}} & \multicolumn{1}{l}{\textbf{HSR-RR}} \\
          \toprule
    \textbf{AR} & 0.8375 & 0.8338 & 0.8329 & 0.8302 & \textbf{0.8377} & 0.8441 & 0.8431 & 0.8444 & 0.8426 & \textbf{0.8454} & 0.8124 & 0.8129 & 0.8113 & 0.8124 & \textbf{0.8152} \\
    \textbf{CR} & 0.7800 & 0.7792 & 0.7718 & 0.7726 & \textbf{0.7824} & \textbf{0.7829} & 0.7800 & 0.7808 & 0.7819 & 0.7792 & 0.7657 & 0.7649 & 0.7628 & 0.7644 & \textbf{0.7673} \\
    \textbf{IMDB} & 0.8392 & 0.8369 & 0.8370 & 0.8281 & \textbf{0.8434} & 0.8491 & 0.8453 & \textbf{0.8493} & 0.8459 & \textbf{0.8493} & 0.7957 & 0.7960 & 0.7953 & 0.7938 & \textbf{0.7999} \\
    \textbf{STS-B} & \textbf{0.8071} & 0.8062 & 0.8048 & 0.8052 & 0.8056 & 0.8044 & 0.8045 & 0.8049 & 0.8031 & \textbf{0.8053} & 0.7818 & 0.7819 & 0.7778 & 0.7813 & \textbf{0.7846} \\
    \bottomrule
    \end{tabular}%
    }
  \label{tab:sentiment}%
\end{table*}%

\section{Experiments}

We evaluate the HSR postprocessing algorithm described in Algorithm \ref{alg:hsr} (denoted by HSR-RR as it is based on ridge regression). We test it on three different pre-trained English word embeddings including Word2Vec\footnote{\url{https://code.google.com/archive/p/word2vec/}} \citep{Mikolov2013}, GloVe\footnote{\url{https://nlp.stanford.edu/projects/glove/}} \citep{Pennington2014}, and Paragram\footnote{\url{https://www.cs.cmu.edu/~jwieting/}} \citep{wieting2015paraphrase}. The original word vectors, as well as word vectors postprocessed by ABTT \citep{Mu2018}, CN \citep{Liu2019}, and SB \citep{Tang2019}, are set as baselines. The performances of these baselines against HSR-RR are compared on word similarity tasks, semantic textual similarity tasks, and downstream sentiment analysis tasks. A statistical significance test is conducted on all experimental results to verify whether our method yields significantly better results compared to the baselines. For ABTT, we set $d = 2$ for GloVe and $d = 3$ for Word2Vec and Paragram as suggested by the original authors. For CN, we fix $d = 2$ for all word embeddings as suggested by the original authors. For HSR, we fix the regularization constants $\alpha_1, \alpha_2 = 50$ for HSR-RR. Generally, we recommend using $\alpha_1, \alpha_2 = 50$ for HSR-RR and other HSR models. Furthermore, we construct a Multilayer Perceptrons HSR model (denoted by HSR-MLP), and the experimental result of HSR-MLP is shown in the appendix.

\subsection{Word Similarity} 

We use seven popular word similarity tasks to evaluate the proposed postprocessing method. The seven tasks are RG65 \citep{Rubenstein1965}, WordSim-353 \citep{Finkelstein2002}, Rare-words \citep{Luong2013}, MEN \citep{Bruni2014}, MTurk \citep{Radinsky2011}, SimLex-999 \citep{Hill2015}, and SimVerb-3500 \citep{Gerz2016}.

For each task, we calculate the cosine similarity between the vector representation of two words, and the Spearman's rank correlation coefficient \citep{Myers1995} of the estimated rankings against the human rankings is reported in Table \ref{tab:word_sim}. In the table, the result marked in bold is the best, and the results underlined are the top three results.

From the table, we could see that while no postprocessing method performs dominantly better than others, HSR-RR has the best performance overall by performing the best on the most number of tasks for two out of the three word embeddings, which are Word2Vec and Paragram. HSR-RR generally achieves the best on these five tasks: WordSim-353, MEN, MTurk, SimLex-999, and SimVerb-3500. Notably, HSR-RR has the best performance on the task SimVerb-3500 for all three word embeddings, which achieves 8.72\%, 40.04\%, and 1.98\% improvement respectively on SimVerb-3500 dataset relative to the original word embeddings and 2.84\%, 9.58\%, and 1.04\% increase compared to the runner-up method. Since SimVerb-3500 is the state-of-the-art task that contains the highest number of word pairs and distinguishes genuine word similarity from conceptual association \citep{Hill2015}, the result obtained on SimVerb-3500 is usually deemed to be more telling than those of other tasks \citep{Liu2019}.

\subsection{Semantic Textual Similarity} 

Next, we test the sentence-level effectiveness of our proposed HSR method on semantic textual similarity (STS) tasks, which measure the degree of semantic equivalence between two texts \citep{Agirre2012}. The STS tasks we employ include 20 tasks from 2012 SemEval Semantic Related task (SICK) and SemEval STS tasks from 2012 to 2015 \citep{Marco2014,Agirre2012,Agirre2013,Agirre2014,Agirre2015}.

To construct the embedding of each sentence in the tasks, we first tokenize the sentence into a list of words, then average the word embedding of all words in the list as the vector representation of the sentence. Following \citet{Agirre2012}, we calculate the cosine distance between the two sentence embeddings and record the Pearson correlation coefficient of the estimated rankings of sentence similarity against the human rankings.

In Table \ref{tab:STS}, we present the result of the 20 STS tasks as well as the average result each year. From the table, we could observe that HSR-RR dominantly outperforms the original word embedding as well as other postprocessing methods, as the average result by year of HSR-RR is the best for all tasks except the SICK task on Word2Vec. On average, HSR-RR improves the Pearson correlation coefficient by 4.71\%, 7.54\%, and 6.54\% respectively over the 20 STS tasks compared to the previously best results, and it achieves 7.13\%, 22.06\%, and 9.83\% improvement respectively compared to the original word embeddings.

\begin{table*}[htbp]
  \centering
  \caption{P-value of one-tailed Student’s t-test of three experiments}
\scalebox{0.77}{%
            \begin{tabular}{lrrrrrrrrrrrr}
            \toprule
          & \multicolumn{4}{c}{\textbf{Word Similarity}} & \multicolumn{4}{c}{\textbf{Semantic Textual Similarity}} & \multicolumn{4}{c}{\textbf{Sentiment Analysis}} \\
           \cmidrule(r){2-5} \cmidrule(r){6-9} \cmidrule(r){10-13}
          & \multicolumn{1}{l}{\textbf{Orig.}} & \multicolumn{1}{l}{\textbf{ABTT}} & \multicolumn{1}{l}{\textbf{CN}} & \multicolumn{1}{l}{\textbf{SB}} & \multicolumn{1}{l}{\textbf{Orig.}} & \multicolumn{1}{l}{\textbf{ABTT}} & \multicolumn{1}{l}{\textbf{CN}} & \multicolumn{1}{l}{\textbf{SB}} & \multicolumn{1}{l}{\textbf{Orig.}} & \multicolumn{1}{l}{\textbf{ABTT}} & \multicolumn{1}{l}{\textbf{CN}} & \multicolumn{1}{l}{\textbf{SB}} \\
          \toprule
    \textbf{WORD2VEC} & \textbf{2.51e-02} & 3.56e-01 & 3.29e-01 & 3.38e-01 & \textbf{2.92e-03} & \textbf{1.12e-03} & \textbf{2.22e-03} & \textbf{1.42e-03} & 9.27e-02 & \textbf{1.35e-03} & \textbf{3.84e-03} & \textbf{2.49e-04} \\
    \textbf{GLOVE} & \textbf{6.85e-03} & 1.83e-01 & 2.30e-01 & \textbf{7.02e-03} & \textbf{2.88e-05} & \textbf{1.35e-03} & \textbf{5.49e-04} & \textbf{5.51e-06} & 4.02e-01 & 4.58e-01 & 1.25e-01 & 1.28e-01 \\
    \textbf{PARAGRAM} & \textbf{4.86e-03} & 7.13e-02 & \textbf{1.62e-02} & 5.13e-02 & \textbf{5.35e-07} & \textbf{1.17e-07} & \textbf{5.94e-07} & \textbf{3.69e-07} & \textbf{1.23e-04} & \textbf{3.32e-04} & \textbf{5.62e-04} & \textbf{1.20e-05} \\
    \bottomrule
    \end{tabular}%

    }
  \label{tab:t_test}%
\end{table*}%

\subsection{Downstream Task: Sentiment Analysis} 

Since the success of intrinsic lexical evaluation tasks does not imply success on downstream tasks, we test the performance of HSR on four sentiment analysis tasks. The dataset we adopt include Amazon reviews\footnote{\url{https://www.kaggle.com/bittlingmayer/amazonreviews\#train.ft.txt.bz2}} (AR), customer reviews (CR) \citep{hu2004mining}, IMDB movie reviews (IMDB) \citep{maas2011learning}, and SST binary sentiment classification (SST-B) \citep{socher2013recursive}, which are all binary sentence-level sentiment classification tasks. Sentiment analysis is an important task in NLP which has been widely applied in business areas such as e-commerce and customer service. 

Similar to the STS tasks, we first tokenize the sentence, then average the corresponding word embeddings as the vector representation of the sentence. We use a logistic regression model trained by minimizing cross-entropy loss to classify the sentence embeddings into positive or negative emotions. This procedure was adopted in previous studies such as \citet{zeng2017socialized}. We report the five-fold cross-validation accuracy of the sentiment classification results in Table \ref{tab:sentiment}.

From Table \ref{tab:sentiment}, we could observe that HSR-RR has the best downstream-task performance among all the tested postprocessing methods. Specifically, for Paragram, HSR-RR achieves the highest classification accuracy on all four tasks; for Word2Vec and GloVe, HSR-RR performs the best on three out of the four tasks.

\subsection{Statistical Significance Test}

To show that our proposed method yields significant improvement compared to the baselines, we employ the one-tailed Student’s t-test. The p-value of the t-test of HSR-RR against other methods for all three experiments is shown in Table \ref{tab:t_test} in scientific notation. We use the convention that a p-value is significant if it is smaller than 0.05, and all significant p-values are marked in bold.

From Table \ref{tab:t_test}, we observe that on word similarity and STS tasks, the improvements yielded by HSR are significant when compared to all three original word vectors. On sentiment analysis tasks, the improvement on Paragram is significant. We also test the significance of improvement of results yielded by HSR-RR with those yielded by other state-of-the-art baseline methods (ABTT, CN, and SB). We find that, for STS tasks, improvements against all three baseline methods on all three word vectors are significant; for sentiment analysis, the improvements against all three baseline methods on Word2Vec and Paragram are significant; for word similarity, only two results (SB on GloVe and CN on Paragram) are significant. While in other cases, improvements of HSR-RR over the original word vectors and the baseline algorithms are not significant, conversely, the baseline methods and the original word vectors also fail to surpass the performance of HSR-RR when the null hypothesis and alternative hypothesis are switched. Therefore, we conclude that HSR-RR yields solid improvement when compared to the original word vectors, and it is either significantly better or on-par with other state-of-the-art baseline methods.

We want to remark that, while statistical significance tests are useful for algorithm comparison, it is mostly excluded in previous word vector evaluation papers \citep{Bullinaria2007,Levy2015,Faruqui2015,Fried2015,Mrksic2016,Mrksic2017,Shiue2017,Mu2018,Liu2019,tang2014learning}, and there could be a valid reason for this. As pointed out by \citet{dror2018hitchhiker}, the way how existing NLP datasets are structured tends to cripple those widely adopted significance tests: while most statistical significance tests (e.g., t-test) assume that the test set consists of independent observations, NLP datasets usually violate this hypothesis. For instance, most STS datasets only contain sentences from a certain source (e.g., news or image captions) and word similarity datasets usually contain words of specialized types (e.g., verbs). This makes a proper significance test quite challenging. Some NLP researchers even contend to abandon the null hypothesis statistical significance test approach due to this hard-to-meet assumption \citep{koplenig2019against,mcshane2019abandon}.

\section{Conclusion and Future Work}

In this paper, we present a simple, fast-to-compute, and effective framework for postprocessing word embeddings, which is inspired by the recent development of causal inference. Specifically, we employ Half-Sibling Regression to remove confounding noise contained in word vectors and to reconstruct latent, ``clean'' word vectors of interest. The key enabler of the proposed Half-Sibling Regression is the linguistic fact that function words and content words are lexically irrelevant to each other, making them natural ``half-siblings''. The experimental results on both intrinsic lexical evaluation tasks and downstream sentiment analysis tasks reveal that the proposed method efficiently eliminates noise and improves performance over the existing alternative methods on three different brands of word embeddings.

The current work has a few limitations, which we wish to address in the future. The first limitation resides in the way we formulate the regression. Note that, when performing the multiple-output regression step in HSR algorithm (Step 1.1 and Step 2.1 of Algorithm \ref{alg:hsr}), we do not take the correlation of targets into account. Such correlations, however, could be beneficial in some cases. Consider, for instance, the task of predicting content words based on stop words (Step 1.1 of Algorithm \ref{alg:hsr}). As content words as targets are strongly correlated (e.g., synonyms and antonyms), such correlations can be further employed to facilitate the regression with well-studied methods such as Reduced-rank regression \citep{Anderson1949}. For a survey of these multiple outcome regression methods taking output into account, please see \citet{Hastie2001}, Section 3.7.

The second line of future work concerns how to use a non-linear model for HSR more effectively. Although we have tried neural-network-based HSR algorithms for various tasks (see the appendix for details), empirically they bring marginally improved results, if not slightly worsened. One hypothesis for explaining this phenomenon is that neural networks tend to be highly expressive, overfitting small datasets easily. For future work, we plan to explore more regularization methods which may improve the results of neural-network-based HSR. 

The third line of future work is to develop a unified framework for understanding word vector postprocessing. As various word vector postprocessing algorithms yield (sometimes surprisingly) similar results in a few cases, it is our hope to establish connections between these approaches in the future. The recent work by \citet{zhou2019getting} points toward this direction.

Last but not least, we believe that there remain ample opportunities for using HSR in other NLP tasks and models. For instance, recently, we have observed that pre-trained language models such as BERT \citep{devlin2019bert} start to replace word vectors as default feature representations for downstream NLP tasks. The HSR framework, in principle, can be incorporated in language model postprocessing pipelines as well. We would like to explore these possibilities in the future.

\paragraph{Acknowledgement} This work was partially supported by the National Natural Science Foundation of China (grant number 71874197). We appreciate the anonymous reviewers for their detailed and constructive comments. We thank all the people who helped Zekun Yang flee from Hong Kong to Shenzhen on Nov. 12th, 2019 such that she could safely finish writing the camera-ready version of this paper.

\bibliography{3106.YangZ.bib}
\bibliographystyle{aaai}

\includepdf[pages={1,2}]{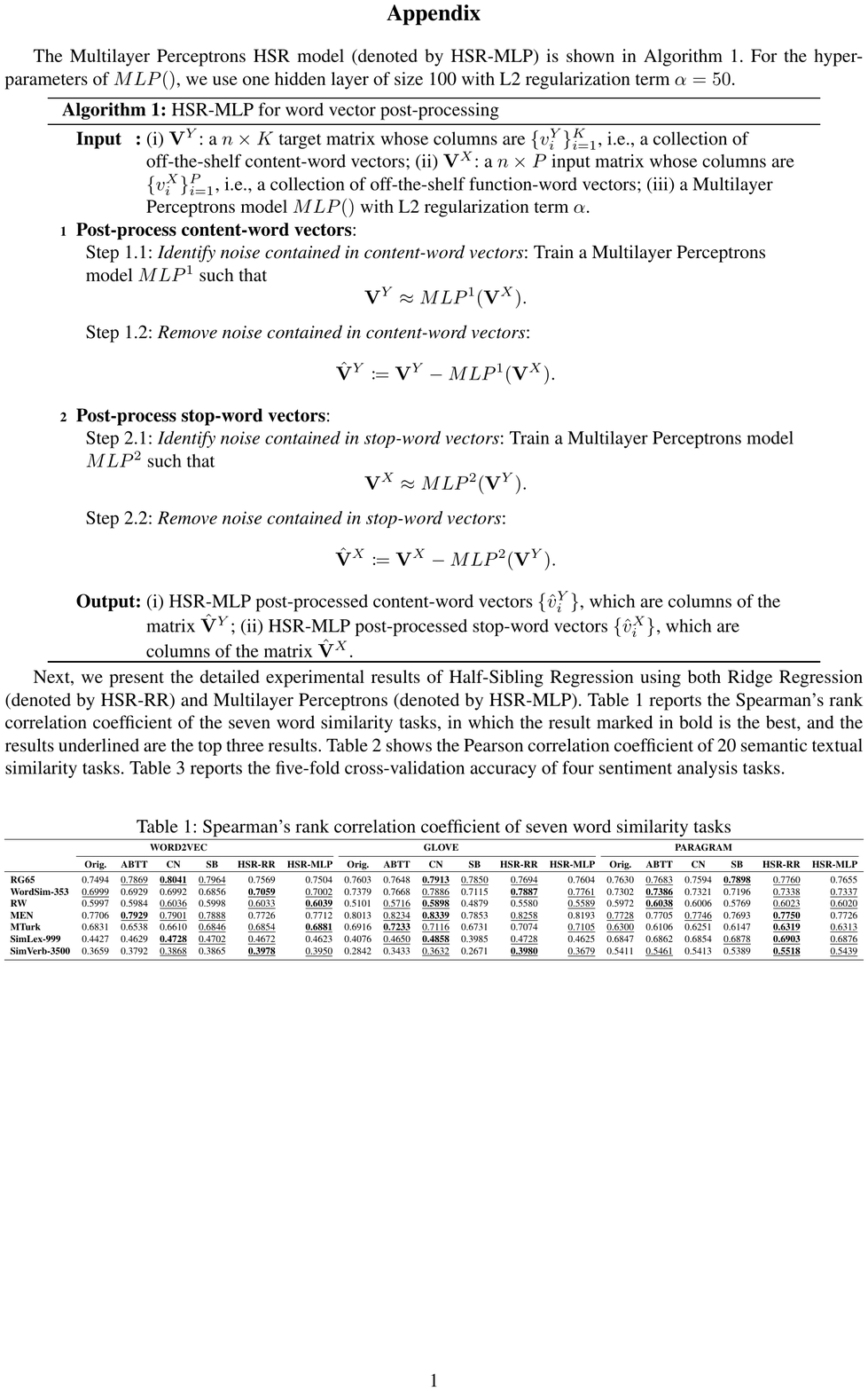}

\end{document}